\documentclass[10pt,twocolumn,letterpaper]{article}
\usepackage[pagenumbers]{cvpr}

\usepackage{amsfonts}
\usepackage{amsmath}
\usepackage{amssymb}
\usepackage{bbm}
\usepackage{tabularx}
\usepackage{tabularray}
\usepackage{enumitem}
\usepackage{graphicx}
\usepackage{xcolor}
\usepackage{lipsum}
\usepackage{wrapfig}
\usepackage{pifont}
\newcommand{\cmark}{\ding{51}}%
\newcommand{\xmark}{\ding{55}}%

\definecolor{cvprblue}{rgb}{0.21,0.49,0.74}
\usepackage[pagebackref,breaklinks,colorlinks,citecolor=cvprblue]{hyperref}

\def\paperID{913} 
\def\confName{CVPR}
\def\confYear{2024}

\title{Latency-aware Road Anomaly Segmentation in Videos: \\A Photorealistic Dataset and New Metrics}
\author{
Beiwen Tian\textsuperscript{1},
Huan-ang Gao\textsuperscript{1},
Leiyao Cui\textsuperscript{2},
Yupeng Zheng\textsuperscript{3},
LUO Lan\textsuperscript{4},
Baofeng Wang\textsuperscript{5},\\
Rong Zhi\textsuperscript{5}, 
Guyue Zhou\textsuperscript{1},
Hao Zhao\textsuperscript{1}\\
\textsuperscript{1}AIR, Tsinghua University,
\textsuperscript{2}Beijing Institute of Technology, 
\textsuperscript{3}Chinese Academy of Science,\\
\textsuperscript{4}Hong Kong Polytechnic University,
\textsuperscript{5}Mercedes-Benz Group China Ltd.\\
{\tt\small tbw23@mails.tsinghua.edu.cn, gha20@mails.tsinghua.edu.cn,}\\ {\tt\small cuileiyaony@gmail.com, zhengyupeng18@mails.ucas.ac.cn,} \\
{\tt\small laura-lan.luo@connect.polyu.hk, baofeng.wang@mercedes-benz.com,}\\
{\tt\small rong.zhi@mercedes-benz.com, zhouguyue@air.tsinghua.edu.cn,} \\
{\tt\small zhaohao@air.tsinghua.edu.cn}
}

\begin{document}
\maketitle

\begin{abstract}
  In the past several years, road anomaly segmentation is actively explored in the academia and drawing growing attention in the industry.
  The rationale behind is straightforward: if the autonomous car can brake before hitting an anomalous object, safety is promoted.
  However, this rationale naturally calls for a temporally informed setting while existing methods and benchmarks are designed in an unrealistic frame-wise manner.
  To bridge this gap, we contribute the \textbf{first video anomaly segmentation} dataset for autonomous driving.
  Since placing various anomalous objects on busy roads and annotating them in every frame are dangerous and expensive, we resort to synthetic data.
  To improve the relevance of this synthetic dataset to real-world applications, we train a generative adversarial network conditioned on rendering G-buffers for photorealism enhancement.
  Our dataset consists of 120,000 high-resolution frames at a 60 FPS framerate, as recorded in 7 different towns.
  As an initial benchmarking, we provide baselines using latest supervised and unsupervised road anomaly segmentation methods.
  Apart from conventional ones, we focus on two new metrics: temporal consistency and \textbf{latency-aware streaming accuracy}.
  We believe the latter is valuable as it measures whether an anomaly segmentation algorithm can truly prevent a car from crashing in a temporally informed setting.
\end{abstract}

\section{Introduction}
With the thriving of autonomous cars, driving safety has become a major concern in the industry.
Despite that substantial progress has been made in the field of visual perception (e.g., semantic segmentation\cite{wang2023one}\cite{wang2023internimage}\cite{fang2023eva}, object detection\cite{li2022yolov6}\cite{li2022efficient}), one of the highest risks of current practices for autonomous driving lies in the fact that uncommon traffic agents cannot be well understood by perception algorithms used by the autonomous driving systems, in which case human driver may fail to response in time and cause catastrophic consequences.
To address the issue, the task of road anomaly segmentation is studied to segment anomalous objects (i.e. objects that are non-existent in training data) in driving scenes where common visual perception models may generate unreliable or confusing predictions.
Timely and accurate segmentation of anomalous objects help determine the timing for human intervention and the guidance for human attention, thus improving the safety of autonomous driving.

\begin{figure*}[htb]
  \centering
  \makebox[\textwidth][c]{\includegraphics[width=1\textwidth]{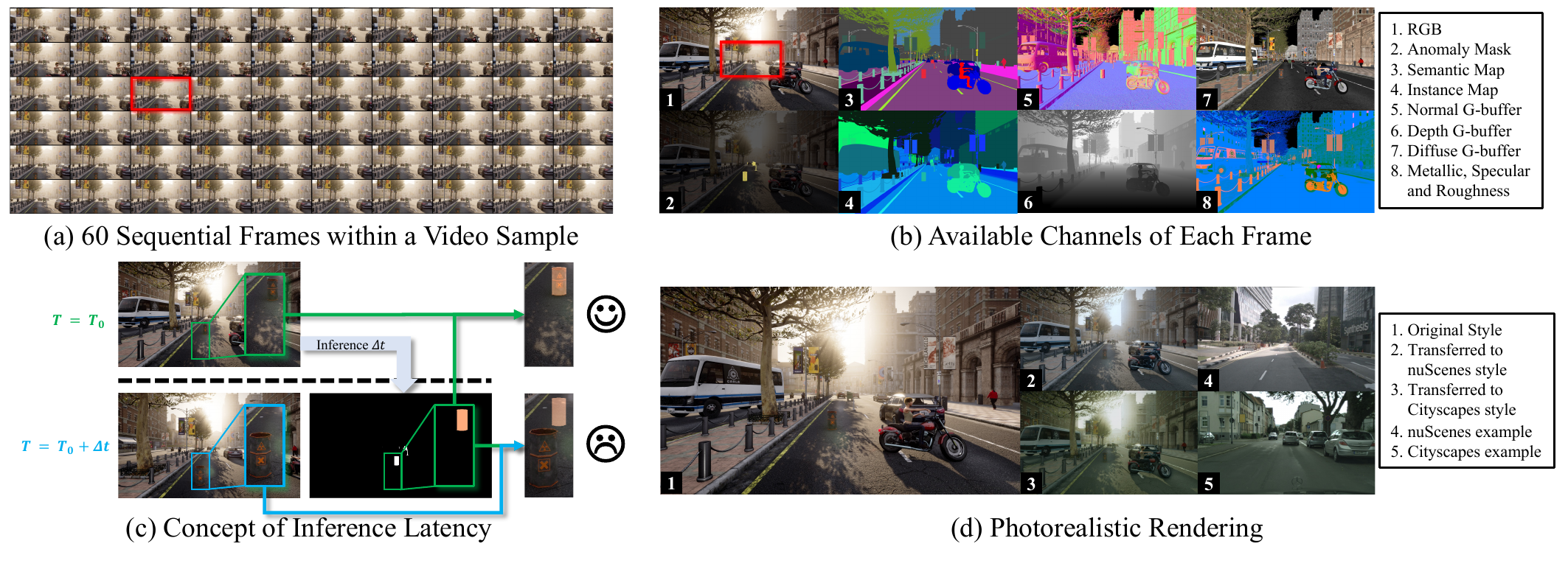}}
  \caption{
    The demonstration of the key features for the proposed benchmark.
    (a) 60 of the total 600 frames in one video sample. Each frame has a resolution of 1920$\times$ 1080 and each video sample has a frame rate of 60 FPS.
    (b) Each frame is recorded with aligned semantic, instance and anomaly map as well as the rendering G-buffers: depth, diffuse, normal, metallic, specular and roughness.
    (c) The concept demonstration for inference latency (i.e., the inference time of the road anoamly segmentation method for evaluation). The predictions of high-latency methods become invalid and impossible for practical uses.
    (d) The comparison before and after applying photorealistic rendering to transfer to the styles of Cityscapes \cite{cordts2016cityscapes} and nuScenes \cite{nuscenes2019}.
  }
  \label{fig:teaser}
\end{figure*}

Recently, academia has seen numerous works to address the task of road anomaly segmentation by uncertainty estimation \cite{lakshminarayanan_simple_2017,kendall_what_2017,oberdiek_detection_2020,lee_training_nodate}, outlier exposure \cite{hendrycks_deep_nodate,hendrycks2022scaling} and image re-synthesis \cite{creusot_real-time_2015,lis2019detecting,xia_synthesize_2020}.
Meanwhile, endeavors \cite{blum2019fishyscapes, Chan2021segmentmeifyoucan, pinggera2016lost, lis2019detecting, hendrycks2022scaling} have also created datasets and benchmarks for the evaluation for road anomaly segmentation.
Still, some drawbacks exist: a) Due to safety concerns, some datasets place the anomalous objects in uncrowded real-world environments (e.g., rural areas or highways), which are not representative of modern driving scenes.
b) Some other datasets create anomalous objects by pasting objects from other domains on existing road scene images, which are easy to identify by domain discrepancy.
c) Mask-form annotations of anomaly regions are scarce, due to the extensive high cost of annotation.
d) The performances of the methods are reported on a per-frame basis, which is inadequate for the evaluation of road anomaly segmentation methods tailored for use in time-sensitive settings.

To address these issues, we resort to synthetic data and contribute the first video anomaly segmentation dataset for autonomous driving.
Our dataset contains 120,000 frames recorded in 7 different towns covering urban and rural areas in CARLA simulator \cite{dosovitskiy_carla_17}, with 21 types of anomalous objects placed ahead of ego vehicle.
The dataset is organized as 200 video sequences, each of which has a length of 10 seconds and a frame rate of 60 FPS (see Fig.~\ref{fig:teaser}(a)).
Each frame has a high resolution of 1920 $\times$ 1080 and is recorded with the ground-truth semantic, instance and anomaly maps with pixel-level precision.
The aligned rendering G-buffers (depth, normal, diffuse, metallic, specular and roughness maps) are also recorded for the use of photorealistic rendering (detailed later).
Available channels of each frame are visualized in Fig.~\ref{fig:teaser}(b).
Dataset details are elaborated in Sec.~\ref{subsec:Motivation} and Sec.~\ref{subsec:Dataset}.

Regarding the metric design for the proposed datasets, we emphasize the importance of the low latency of road anomaly segmentation methods.
Following \cite{li_towards_2020}, the latency is defined as the inference time of a given method during which the input image from the environment may have changed.
The predictions by high-latency methods, even the accurate ones, may become invalid as the ego vehicle may have advanced a considerable distance within the high latency of the method (see Fig.~\ref{fig:teaser}(c)).
To address the lack of evaluation for latency in road anomaly segmentation, we propose a new set of benchmark metrics including latency-aware streaming AUROC and FPR95, which apply to sequences of frames.
With these metrics, the ground-truth anomaly masks of future frames are selected for the evaluation of anomaly segmentation, reflecting both the performance and the latency of methods for evaluation.
Metric designs are detailed in Sec.~\ref{subsec:Benchmark Metrics}.


With a high frame rate of 60 FPS, our proposed dataset provides a naturally fine-grained measure for the latency and is suitable for the proposed metrics.
Still, due to the domain gap between synthetic and real driving scenes, it may be difficult for methods with outstanding performances on our proposed benchmark to be directly applied in reality.
Therefore, alongside the proposed dataset, we follow Richter \etal \cite{richter_enhancing_2022} and provide a toolchain for photorealistic rendering which is capable of transferring any frame to the style of given driving scenes (e.g. cityscapes, nuScenes or any captured road images).
The recorded rendering G-buffers (see Fig.~\ref{fig:teaser}(b)) serve as additional guidance for photorealistic rendering.
The transferred results with the Cityscapes \cite{cordts2016cityscapes} and nuScenes \cite{nuscenes2019} styles are depicted in Fig.~\ref{fig:teaser}(d), while details are elaborated in Sec.~\ref{subsec:Enhancement}.
We conduct benchmark experiments on the original style as well as the transferred styles and report results in Sec.~\ref{sec:Benchmark Results}.







In summary, our contributions in this work are three-fold:
\begin{enumerate}
  \item We contribute the first video dataset for road anomaly segmentation. Video sequences are recorded with high resolution, high frame rate and multiple channels.
  \item We propose the innovative latency-aware metrics for the benchmark of anomaly segmentation regarding both the accuracy and streaming latency.
  \item We provide a photorealistic rendering toolkit for the dataset to transfer to any style of existing driving scenes.
\end{enumerate}
Codes, data and models will be publicly available.

\section{Related Works}
\paragraph{Datasets for road anomaly}
Road anomaly is a serious threat to the safety of autonomous driving.
However, few existing datasets focus on the very task of road anomaly detection or segmentation.
Road Anomaly dataset\cite{lis2019detecting} contains 60 images of unusual dangers on the road which are annotated in a pixel-wise manner.
LostAndFound\cite{pinggera2016lost} is a dataset for road obstacle segmentation with more than two thousand frames and pixel-wise annotations of obstacles on the road.
Fishyscapes\cite{blum2019fishyscapes} is a public dataset for the task of road anomaly segmentation, with three splits \textit{Web}, \textit{Static} and \textit{Lost \& Found}\cite{pinggera2016lost}.
The first two create scenes with anomalies by pasting existing or web-crawled object images in the road scenes, and the third is based on the LostAndFound dataset.
Road Anomaly Detection Dataset (RADD)\cite{li_towards_2020} is a video dataset for road anomaly detection that has 1,000 video clips of 10 seconds each with 500 of them contains anomalies.
In the field of road anomaly detection and segmention, datasets are mostly based on images, and video-based datasets are relatively less common.
Meanwhile, no video-based datasets are intended for the task of anomaly segmentation.
The datasets proposed in this work fill this vacancy.

\paragraph{Synthetic road scene datasets}
With the development of rendering technologies, simulated scenes have become increasingly realistic and many simulated datasets have been proposed.
As for the road scenes, PfD\cite{richter2016playing}, PfB\cite{richter2017playing}, SYNTHIA\cite{ros2016synthia} and Virtual KITTI\cite{gaidon2016virtual} are collected in simulators or game engines and proposed for the task of road scene understanding.
Of all the datasets, the MUAD dataset\cite{franchi2022muad} is the first synthetic dataset for road anomaly detection.
The dataset contains 10413 annotated frames in total, with 1668 containing anomalous (OOD) objects.
Few of existing synthetic road scene datasets focus on the task of anomaly segmentation, and none of them is video-based.
What's more, the domain gap between synthetic and real road scenes is still a challenge for the application of the existing synthetic datasets in real-world scenarios.
The enhancement toolkit provided in this work is able to fill the domain gap to some extent and make the out synthetic dataset more applicable to transfer to real-world scenarios.



\section{Benchmark Design}
\label{sec:Benchmark Design}

\subsection{Motivation}
\label{subsec:Motivation}

Semantic segmentation serves as the key role of understanding surrounding environments in the autonomous driving systems.
Yet, widely-used algorithms (e.g. InternImage\cite{wang2023internimage}, Vit-Adapters\cite{chen2022vision}) used for the very task is trained on a pre-defined set of categories (e.g. the 19 categories in Cityscapes dataset\cite{cordts2016cityscapes}) which is incapable of handling the novel objects in the real world.
Anomaly segmentation, therefore, is a necessary functionality in autonomouse driving systems by indicating when and where the semantic segmentation fails to understand correctly and human intervention is needed.

It is worth noting that, both the accuracy and the inference time (which we define as \textbf{latency}) of the anomaly segmentation methods should be investigated before being used as guidance for human intervention.
The high accuracy of anomaly segmentation methods reflects precise localization of the anomalous objects in road scenes, which leads to more focused attention and lower response time for the human driver.
The low latency, on the other side, is another key factor in the application of anomaly segmentation, as demonstrated in Fig.~\ref{fig:teaser}(c).
Suppose we use an oracle anomaly segmentation method which produces correct anomaly masks with high latency in the autonomous driving application.
Despite that the method produces ideal anomaly masks for the input frame at time $T_0$, the ego vehicle will have advanced a significant distance during the high latency $\Delta T$, which makes the anomaly masks for $T_0$ invalid for the input frame at $T_0 + \Delta T$.
The mismatched predictions could result in catastrophic consequences as human attention is guided to focus on non-anomalous regions.
Unfortunately, existing benchmarks for anomaly segmentation (e.g., Fishyscapes\cite{blum2019fishyscapes}) only adopt segmentation metrics (e.g., AUROC, AP, FPR95) and overlook the latency during evaluation, which makes the existing benchmarks incomplete for practical tests of anomaly segmentation algorithms.

To address this gap, we contribute the first latency-aware road anomaly segmentation benchmark with a large-scale video dataset and the innovative streaming metrics.
To avoid the enormous efforts for anomaly mask annotation, we resort to CARLA simulator\cite{dosovitskiy_carla_17} and collect a large-scale video dataset with high resolutions and high frame rates, with mid-level rendering results (e.g., depth map, normal map, diffuse map and specular map) included. Details are in Sec.~\ref{subsec:Dataset}.
Additionally, to bridge the gap between simulation and reality, we follow Richter \textit{et al.}\cite{richter_enhancing_2022} and provide a photorealistic rendering toolkit which enables transferring to any style in given road scene images (detailed in Sec.~\ref{subsec:Enhancement}).
Based on the high-framerate videos we collect, we build a benchmark with various metrics including the innovative \textbf{latency-aware streaming metrics} which prefers methods with high accuracy and low latency at the same time (detailed in Sec.~\ref{subsec:Benchmark Metrics}).








\subsection{Dataset}
\label{subsec:Dataset}

\begin{figure*}[htb]
  \centering
  \includegraphics[width=1.0\textwidth]{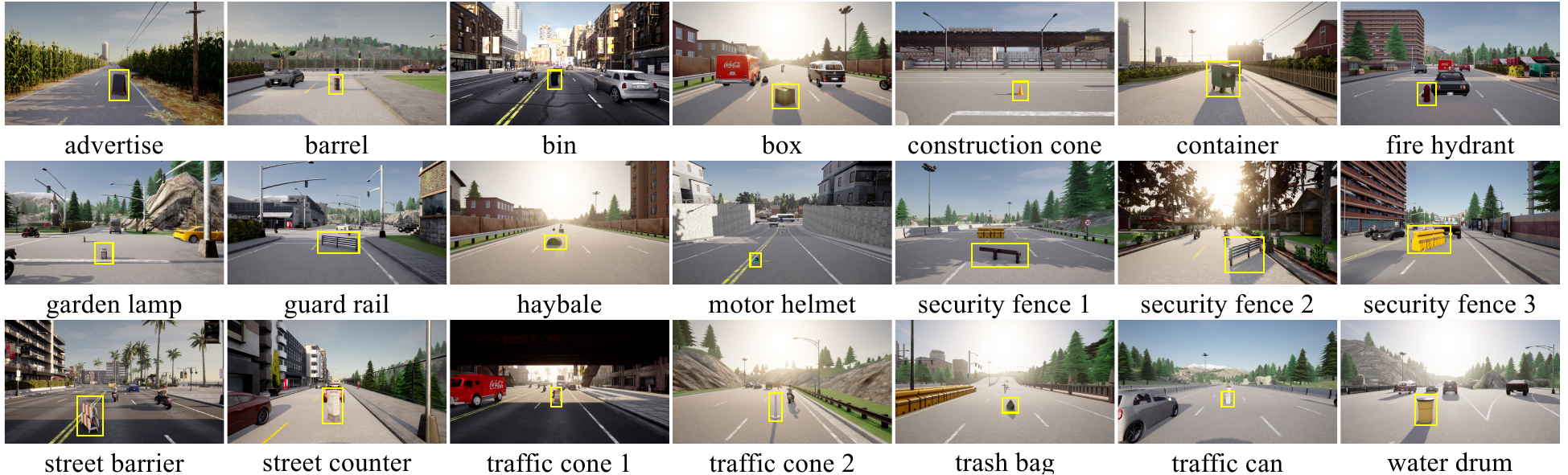}
  \caption{
    The demonstration of available anomalous objects in the dataset.
    Anomalous objects are emphasized by the yellow boxes added in postprocessing.
    }
  \label{fig:anomaly-objects}
\end{figure*}



In this section, we introduce the dataset collected and used for the benchmark.
Despite that various datasets have been proposed for the task of anomaly segmentation (e.g., Fishyscapes\cite{blum2019fishyscapes}, RoadAnomaly\cite{lis2019detecting}, LostAndFound\cite{pinggera2016lost}), existing datasets are collected or annotated on a frame basis and are not suitable for evaluation in the temporal-informed setting.
To adequately incorporate inference latency during evaluation, a new video-based dataset for anomaly segmentation is needed with precise annotations on anomalous objects in each frame.

The ideal way to create the dataset would be capturing video sequences with anomalous objects placed on roads, each frame of which is then annotated by human annotator.
However, this process is not practical since placing anomalous objects in driving lanes would bring significant security risks to driving vehicles and manual pixel-wise anomaly mask annotation would be excessively cost-inefficient.
Some efforts\cite{blum2019fishyscapes} resort to synthetic anomalous scenes by pasting objects from other datasets\cite{everingham2010pascal} onto normal road scenes images, but the pasted objects have significantly different lighting conditions from the background which makes the anomalous objects easier to identify in these synthetic scenes than in reality scenes.

To this end, we resort to simulation for generation of video frames and aligned anomaly masks.
We choose CARLA \cite{dosovitskiy_carla_17} as the simulator for road scenes, as CARLA is released with open digital assets including several manually crafted towns and is capable of rendering with various lighting conditions.

We made certain modifications to CARLA simulator.
1) We add the support for anomaly semantic category, so that the anomalous objects placed in the scene would be identified separately.
2) We add a policy to spawn anomalous objects in front of the driving ego vehicle in the range of 10 to 50 meters.
Anomalous objects are defined as objects that never appear in road lanes in front of the ego vehicle.
3) We add inceptors to record intermediate rendering results (e.g., depth maps, normal maps, diffuse maps and irradiance maps etc., usage detailed later in Sec.~\ref{subsec:Enhancement}) and the extrinsics parameters of the ego vehicles (usage detailed later in Sec.~\ref{subsubsec:Temporal Consistency Metric}).

\begin{figure}
  \includegraphics[width=0.5\textwidth]{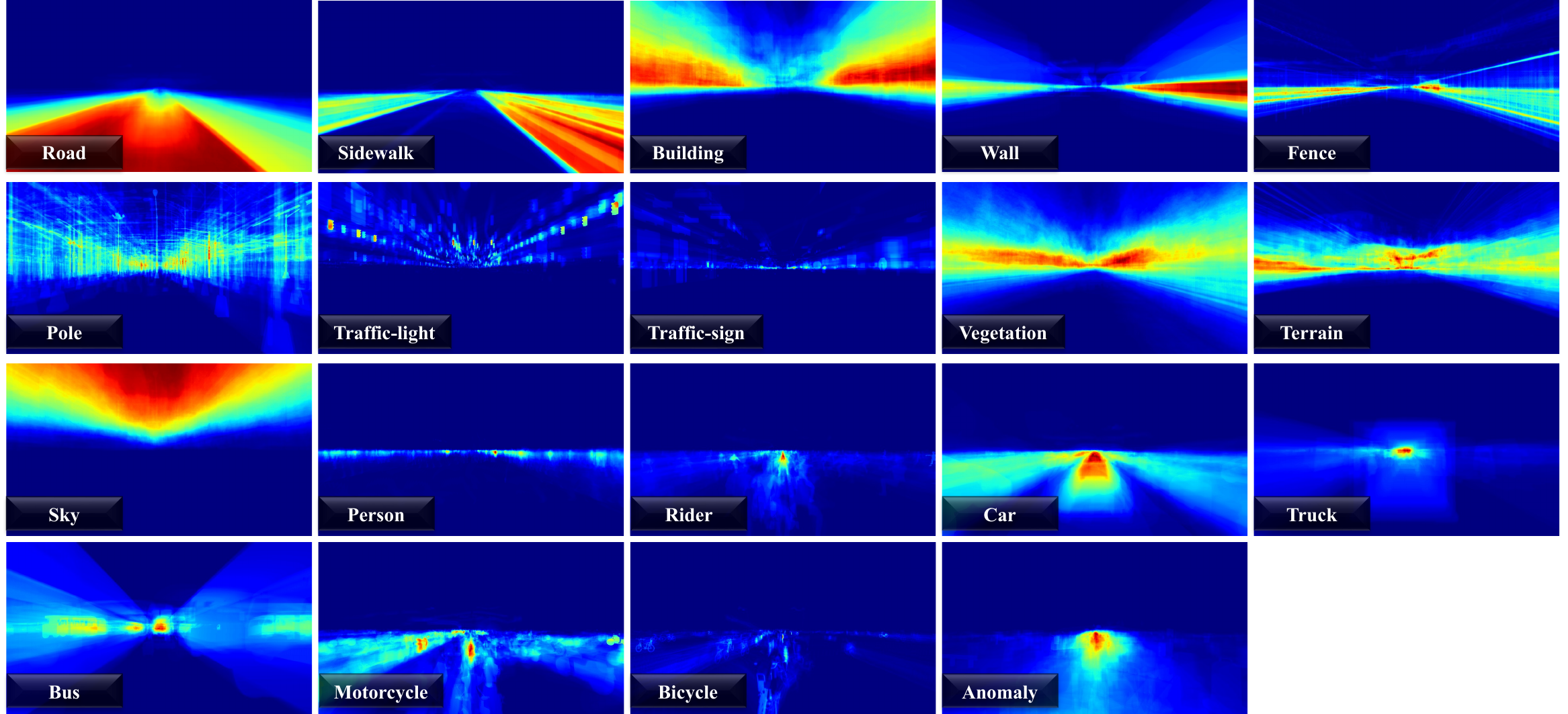}
  \caption{
    The spatial distribution of anomalous objects.
    Red area demonstrates higher frequency of anomalous objects.
  }
  \label{fig:anomaly-distribution}
\end{figure}

With the modify CARLA simulator, we record 220 video sequences at a frame rate of 60 FPS, each of which contains 600 frames.
The first 60 frames of a certain sequence are demonstrated in Fig.~\ref{fig:teaser}(a).
Each frame has a resolution of 1920 x 1080 and is recorded with pixel-level anomaly, semantic and instance maps.
Available channels are shown in Fig.~\ref{fig:teaser}(b).
The label protocol for the semantic maps is the same as Cityscapes\cite{cordts2016cityscapes}, and the available anomalous objects are shown in Fig.~\ref{fig:anomaly-objects}.
We further visualize the spatial distribution of anomalous objects in Fig.~\ref{fig:anomaly-distribution}.

To the best of our knowledge, our dataset is the first video datasets for road anomaly segmentations, not to mention the high resolution and the high frame rate which enables streaming evaluation metrics introduced in Sec.~\ref{subsec:Benchmark Metrics}.

\subsection{Enhancement Toolkit}
\label{subsec:Enhancement}



Although simulated data can be collected with high efficiency, the domain gap between the simulated data and real-world prevents anomaly methods trained on simulated data from transferring to real-world driving scenes.
Furthermore, the real-world driving scenarios are highly diverse due to different lighting conditions and sensor choices, which makes it difficult for anomaly segmentation methods to generalize well.
In order to fill the domain gap and further enable data augmentation under different conditions, we propose to leverage generative models for photorealistic enhancement \cite{richter_enhancing_2022} to transfer the simulated scenes to \textit{arbitrary} realistic scenarios.

More specifically, a GAN-based enhancement framework with a generator and a discriminator is adopted.
The generator is a HRNetV2 \cite{wang2020deep} based encoder-decoder network, which is intended to render the simulated images with realistic styles.
Inspired by the exciting success of EPE\cite{richter_enhancing_2022}, we leverage the deferred shading results (i.e., \textbf{G-buffers}, including ground-truth normal, depth, diffuse, metallic, specular and roughness maps) of the simulated images as the auxiliary inputs.
These aligned G-buffers are extracted during the rendering process of the CARLA simulator with a custom plug-and-play inceptor (illustrated in Fig.~\ref{fig:teaser}(b)).
The use of the G-buffers is to ensure that the generator would generate realistically enhanced images without largely altering geometry and materials of the simulated scenes.
It is worth noting that, with these auxiliary inputs as the controlling conditions, the styles of the frames within a video sequence are natually consistent.

The discriminator, on the other hand, intends to minimize the perceptual likeliness between enhanced images and the real-world scenes (i.e., realism score) as well as the LPIPS distance \cite{zhang2018perceptual} between the enhanced images and the original simulated images.
In the calculation of the realism scores, pixel-wise semantics are required, which we obtain with MSeg \cite{lambert2020mseg}, a pre-trained robust segmentation network for driving scenes.
With MSeg, manual labeling of the real-world scenes is not necessary, which means the enhancement pipeline is available to both existing real-world road scenes datasets (e.g., Cityscapes\cite{cordts2016cityscapes} and nuScene\cite{nuscenes2019}) and any manually captured road scene images.

\begin{figure*}[htb]
  \centering
  \includegraphics[width=1.0\textwidth]{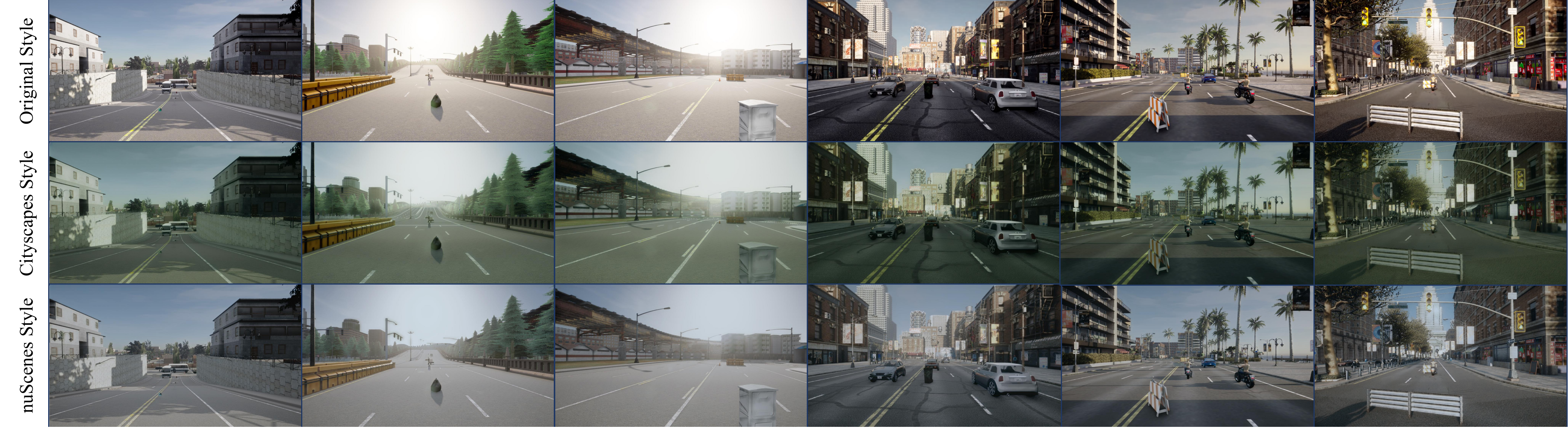}
  \caption{
    The examples of photorealistic enhancement on the collected simulated video frames.
    The anomalous objects are also enhanced with the same style.
    }
  \label{fig:qualitative}
\end{figure*}

Once trained to convergence, the enhancement network is capable of performing data augmentation by enhancing the collected video sequences to resemble the characteristics and complexities of any target environment.
We perform this process on the collected video sequences to obtain video sequences with the styles of Cityscapes\cite{cordts2016cityscapes} and nuScene\cite{nuscenes2019}, which are released alongside the original dataset and also used for benchmark in Sec.~\ref{sec:Benchmark Results}.
Some enhanced examples are illustrated in Fig.~\ref{fig:qualitative}.
It is worth noting that the anomalous objects are also enhanced, resulting in overall more realistic scenes than former arts using cut-and-paste to create scenes with anomalous objects.

Furthermore, the whole enhancement pipeline is organized and also released along with the dataset itself, with which we hope to boost the development of anomaly segmentation algorithms by providing a more realistic large-scale dataset for evaluation.




\subsection{Benchmark Metrics}
\label{subsec:Benchmark Metrics}

In this section, we introduce the metrics we design and use for anomaly segmentation in the temporally informed setting.
The metrics are three-fold: latency-agnostic metrics in Sec.~\ref{subsubsec:Latency-agnostic Metrics}, latency-aware streaming metrics in Sec.~\ref{subsubsec:Latency-aware Streaming Metrics}, and the temporal consistency metric in Sec.~\ref{subsubsec:Temporal Consistency Metric}.

\subsubsection{Latency-agnostic Metrics}
\label{subsubsec:Latency-agnostic Metrics}
We first review the latency-agnostic metrics which evaluate video anomaly segmentation methods on a per-frame basis.
Given a frame $\mathbf{x}_t \in \mathbb{R}^{H \times W \times C}$ in a video $\mathbf{X} \in \mathbb{R}^{T \times H \times W \times C}$, where $T$ denotes the temporal dimension, an anomaly score map $\mathbf{\hat{y}}_t \in \mathbb{R}^{H \times W}$ is predicted by the evaluated method $\phi$, which highlights outlier pixels with higher scores.
The anomaly score map is compared with the ground-truth anomaly label map $\mathbf{y}_t \in \mathbb{R}^{H \times W}$ in which the value $1$ indicates anomaly and the value $0$ indicates normality.

In the context of anomaly segmentation, the final anomalous regions are obtained by selecting a threshold for the predicted anomaly score maps.
The selection can be challenging as it is a tradeoff between false negative predictions and false positive predictions.
To address this issue, area-under metrics are commonly adopted to provide a comprehensive assessment of the model's ability to distinguish anomalies from normal regions across different threshold settings.
Two widely used metrics in this regard are \textbf{AUROC} (Area Under the Receiver Operating Characteristic) and \textbf{AUPRC} (Area Under the Precision-Recall curve).
Additionally, \textbf{FPR@95} (False Positive Rate at a true positive rate threshold of 95$\%$) is often utilized to provide insights into the model's ability to maintain a high detection rate while keeping the false positive rate at a desired level.

To this end, the \textbf{latency-agnostic} performance $f$ of an anomaly segmentation method $\phi$ on a video sequence $\mathbf{X}$ is defined as
\begin{align}
  f(\phi, \mathbf{X}, \mathbf{Y}) = \frac{1}{T} \sum_{t=1}^T \text{M} (\phi(\mathbf{x}_t), \mathbf{y}_t)
\end{align}
where $\text{M} \in \{\text{AUROC, AUPRC, FPR@95}\}$ and $T$ is the number of frames in the video sequence.


\begin{figure*}[htb]
  \centering
  \makebox[\textwidth][c]{\includegraphics[width=0.95\textwidth]{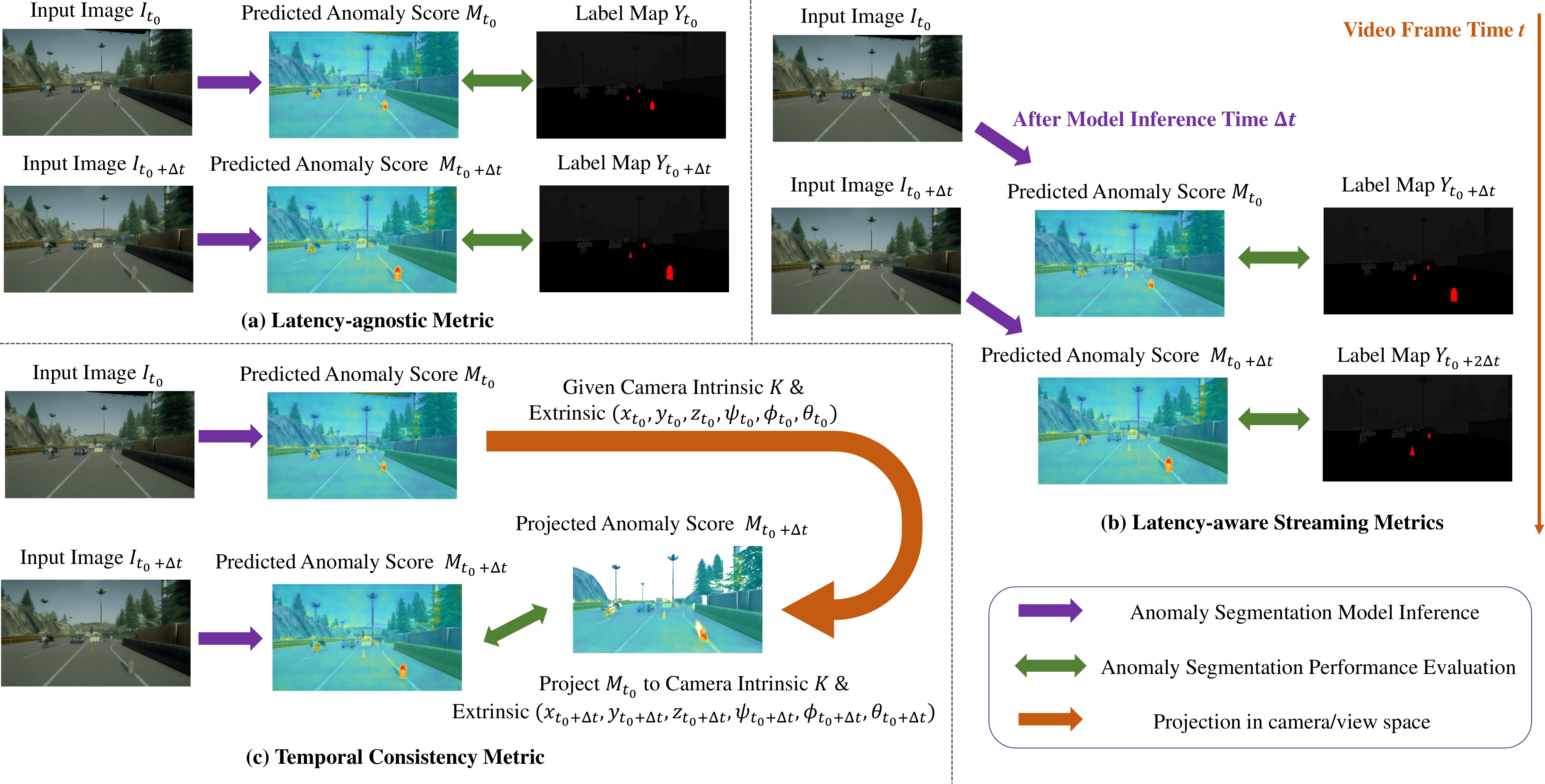}}
  \caption{Illustration of the proposed metrics.
    \textbf{(a)} \textbf{Latency-Agnostic Metrics.}
    The evaluation is performed on a frame basis, irrelevant of time or latency.
    \textbf{(b) Latency-Aware Streaming Metrics.}
    The predicted anomaly score map for the input frame at time $T_0$ is compared with the ground truth anomaly map at time $T_0 + \Delta t$.
    Here $\Delta t$ is the latency of the method.
    \textbf{(c) Temporal Consistency Metric.}
    The predicted anomaly map at time $T_0$ is projected to the image space at $T_0 + \Delta t$ and compared with the predicted anomaly map at time $T_0 + \Delta t$.
    The latency $\Delta t$  between the two demonstrated frames is 60 frames (or equivalently 1 second).
  }
  \label{fig:metric}
\end{figure*}

\subsubsection{Latency-aware Streaming Metrics}
\label{subsubsec:Latency-aware Streaming Metrics}


While the latency-agnostic metrics are well-suited for image anomaly segmentation, they are not optimal for evluation on video sequences as the latency of the evaluated methods is not taken into account.
In reality, a segmentation method with a excessively high latency may still achieve high performance with the latency-agnostic metrics but fail to provide timely feedback for decision-making in the real world.
The phenomenon makes a request for metrics that prefer methods with both high accuracy and lower latency.

To this end, we propose \textbf{latency-aware streaming metrics} that incorporate the latency (i.e., the inference time $\Delta t$ of the anomaly segmentation method $\phi$).
The metrics are motivated by the scenario where a driving vehicle adopts an anomaly segmentation algorithm to detect the anomalous regions on the road and guide the possible human intervention.
As illustrated in Fig.~\ref{fig:teaser}, with a latency of $\Delta t$, the anomalous objects in the input frame at time $T_0$ can only be detected by the algorithm until $T_0 + \Delta t$, at which time the vehicle may have advanced a long distance causing anomalous regions to shift.
From the perspective of the driver, the prediction of frame at $T_0$ is then used for guidance for intervention at $T_0 + \Delta t$.
We intend to mimic this perspective, and propose the latency-aware streaming metrics by evaluating the segmentation results at $T_0$ with the ground truth at $T_0 + \Delta t$. Namely, the \textbf{latency-aware streaming} performance $f$ of an anomaly segmentation method $\phi$ on a video sequence $\mathbf{X}$ is defined as:
\begin{align}
  f(\phi, \mathbf{X}, \mathbf{Y}) = \frac{1}{T - \Delta t} \sum_{t=1}^{T - \Delta t} \text{M}(\phi(\mathbf{x}_t), \mathbf{y}_{t+\Delta t})
a\end{align}
where $\text{M} \in \{\text{AUROC, AUPRC, FPR@95}\}$, $T$ is the number of frames in the video sequence, $\Delta t$ is the latency of the evaluated method $\phi$.
Here, the latency of the method is represented by the number of frames in-between, since the frame rate of the video sequence is fixed at 60 fps.
$\mathbf{x}_t$ and $\mathbf{y}_t$ denote the input frame and the ground truth anomaly map at time $t$, respectively.

\begin{figure}[htb]
  \centering
  \includegraphics[width=0.48\textwidth]{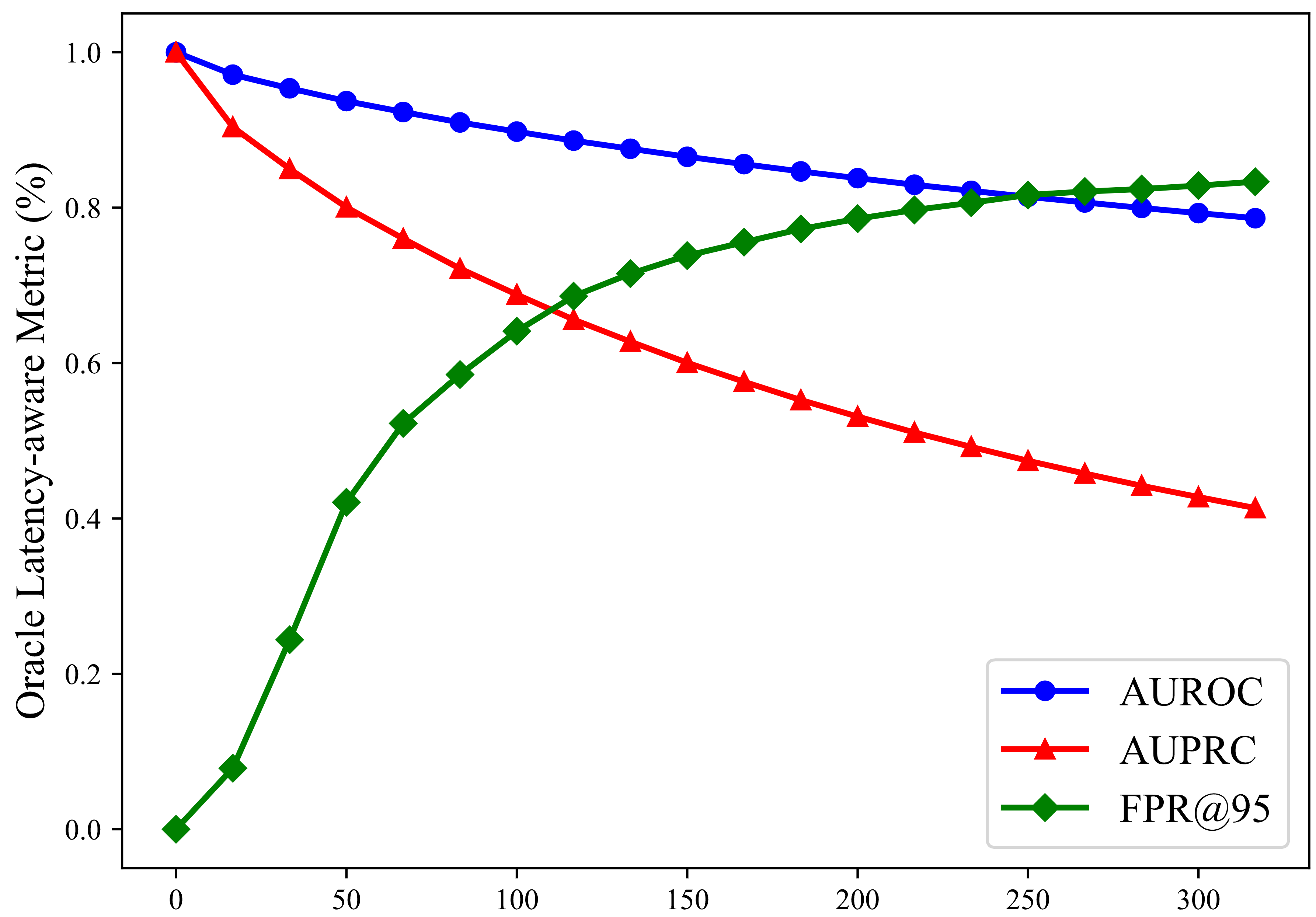}
    \caption{The oracle segmentation result for the video sequence in Fig.~\ref{fig:teaser}(a). The x-axis denotes the manually set latency for the oracle segmentation results in number of frames.}
  \label{fig:oracle}
\end{figure}

\begin{table*}
  \caption{
    The results of recent state-of-the-art methods on the proposed benchmark. The highest score in each transferred style and each metric are \textbf{emphasized}.  \\
  }
  \label{table:main}
  \centering
  \resizebox{0.95\linewidth}{!}{
    \begin{tblr}{
      cells = {c,},
      cell{1}{1-4, 11-12} = {r=2}{},
      cell{1}{5, 8} = {c=3}{},
      hline{1,3}={.8pt},
      cell{3,8,13}{1} = {r=5}{},
      hline{8,13} = {},
          hline{18}={.8pt},
        }
      Style & Method                                  & Extra Data & Retraining & Inference-time-agnostic Metrics &                       &                          & Inference-time-aware Metrics &                       &                          & {Temporal            \\ Consistency ($\%$) $\uparrow$ }& {Frame Inference \\ Time (ms) $\downarrow$} \\
            &                                         &            &            & AUROC $\uparrow$                & \rm{AUPRC} $\uparrow$ & \rm{FPR@95} $\downarrow$ & \rm{AUROC} $\uparrow$        & \rm{AUPRC} $\uparrow$ & \rm{FPR@95} $\downarrow$ &                &     \\

      \rotatebox{90}{Original}
            & SML \cite{jung2021standardized}         & \xmark     & \xmark     & 97.49                           & \textbf{67.35}        & 8.96                     & 97.02                        & \textbf{63.78}        & 11.13                    & \textbf{94.32} & 33  \\
            & GMMSeg \cite{liang2022gmmseg}           & \xmark     & \xmark     & 57.71                           & 1.97                  & 66.17                    & 57.64                        & 1.92                  & 65.83                    & 83.84          & 366 \\
            & PEBAL \cite{tian2022pixel}              & \cmark     & \cmark     & 98.01                           & 52.48                 & 7.10                     & 73.22                        & 16.71                 & 63.67                    & 57.05          & 587 \\
            & RPL \cite{liu2022residual}              & \cmark     & \cmark     & 98.12                           & 64.69                 & 5.04                     & \textbf{97.42}               & 61.79                 & \textbf{9.93}            & 75.20          & 19  \\
            & DenseHybrid \cite{grcic2022densehybrid} & \cmark     & \cmark     & \textbf{98.30}                  & 38.08                 & \textbf{4.93}            & 95.62                        & 33.28                 & 20.73                    & 89.21          & 22  \\

      \rotatebox{90}{Cityscapes}
            & SML \cite{jung2021standardized}         & \xmark     & \xmark     & 95.86                           & 38.36                 & 13.03                    & 95.00                        & 36.21                 & 15.63                    & \textbf{97.64} & 33  \\
            & GMMSeg \cite{liang2022gmmseg}           & \xmark     & \xmark     & 55.16                           & 1.86                  & 67.88                    & 55.14                        & 1.82                  & 67.43                    & 85.70          & 366 \\
            & PEBAL \cite{tian2022pixel}              & \cmark     & \cmark     & 96.63                           & 32.51                 & 11.02                    & 78.47                        & 9.67                  & 45.25                    & 88.12          & 587 \\
            & RPL \cite{liu2022residual}              & \cmark     & \cmark     & 97.83                           & \textbf{70.49}        & \textbf{4.58}            & \textbf{97.29}               & \textbf{67.38}        & \textbf{7.81}            & 84.67          & 19  \\
            & DenseHybrid \cite{grcic2022densehybrid} & \cmark     & \cmark     & \textbf{97.94}                  & 45.02                 & 5.54                     & 95.77                        & 39.85                 & 20.57                    & 78.61          & 22  \\

      \rotatebox{90}{nuScenes}
            & SML \cite{jung2021standardized}         & \xmark     & \xmark     & 95.00                           & 25.80                 & 16.00                    & 93.69                        & 23.12                 & 20.78                    & 86.26          & 33  \\
            & GMMSeg \cite{liang2022gmmseg}           & \xmark     & \xmark     & 57.61                           & 2.07                  & 65.35                    & 57.49                        & 2.03                  & 65.02                    & 85.38          & 366 \\
            & PEBAL \cite{tian2022pixel}              & \cmark     & \cmark     & 81.01                           & 35.88                 & 32.70                    & 81.82                        & 18.85                 & 37.09                    & 85.35          & 587 \\
            & RPL \cite{liu2022residual}              & \cmark     & \cmark     & \textbf{98.38}                  & \textbf{69.37}        & \textbf{3.37}            & \textbf{98.02}               & \textbf{66.85}        & \textbf{5.89}            & \textbf{92.13} & 19  \\
            & DenseHybrid \cite{grcic2022densehybrid} & \cmark     & \cmark     & 95.50                           & 34.02                 & 8.77                     & 94.79                        & 32.93                 & 21.64                    & 77.48          & 22  \\
    \end{tblr}
  }
\end{table*}


It is important to highlight two key remarks regarding these metrics.
\textbf{(a)}
Since the video sequences are discrete, the frame with the closest timestamp to $T + \Delta t$ is used as the ground truth map.
Therefore, time interval between frames, or equivalently the frame rate of the videos, makes a crucial difference as it determines the preciseness of the penalty incurred by the latency.
With a high framerate of 60 FPS, our dataset proposed in Sec.~\ref{subsec:Dataset} is the most suitable among all for evaluating with the proposed latency-aware streaming metrics.
\textbf{(b)} The metrics is also biased by the spatial distribution of anomalies within the scene.
Specifically, the metrics is more affected by objects or regions that are closer to the camera compared to those farther away, since the latter tends to have more stable locations in the frames than the former according to the perspective principle.
This is also consistent with realistic scenarios where the driver should be more concerned about the anomalies that are closer to the ego vehicle.

We further evaluate with the proposed latency-aware metrics on an oracle method that hypothetically achieves absolutely correct anomaly segmentation result with different latencies.
This is implemented by comparing the ground-truth anomaly maps at different timestamps with the latency-aware metrics.
As shown in the Fig.~\ref{fig:oracle}, latency-aware AUROC and AUPRC decreases as the latency increases, while latency-aware FPR@95 increases as the latency increases, as expected.

\subsubsection{Temporal Consistency Metric}
\label{subsubsec:Temporal Consistency Metric}
Another important aspect of video anomaly segmentation is the temporal consistency of the method, without which the method may produce inconsistent results for the same scene with small temporal perturbations, causing confusion to the driver and potentially leading to accidents.
Therefore, we include an extra metric to evaluate the temporal consistency of an anomaly segmentation methods $\phi$ (with a latency of $\Delta t$) in the following steps:

\paragraph{Obtaining the anomaly masks.} Firstly, we determine thresholds for the predicted anomaly scores of $\hat{\mathbf{y}}_t = \phi(\mathbf{x}_t)$ and $\hat{\mathbf{y}}_{t + \Delta t} = \phi(\mathbf{x}_{t + \Delta t})$ respectively based on the True Positive Rate (TPR) at 95\% with the guidance of ground truth labels. With the thresholds, binary anomaly masks $S_t$ and $S_{t + \Delta t}$ are generated with the size of $H \times W$.

\paragraph{Projection.} Using the camera intrinsics $K$, depth maps of the two frames and the relative view transformation matrix obtained from extrinsics of the two frames, we project the predicted anomaly mask $S_t$ into the image space of camera at time $t + \Delta t$ and obtain $S_{t \rightarrow t + \Delta t}$.
This projection is achieved by mapping each pixel in $S_t$ to its corresponding position in the image space of $t + \Delta t$, as illustrated in Fig.~\ref{fig:metric}(c).
To ensure precision, we impose clipping at a maximum depth of 80 meters.

\paragraph{Calculating IoU.} We compare the projected predicted anomaly mask $S_{t \rightarrow t + \Delta t}$ with the predicted anomaly mask $S_{t + \Delta t}$ and report Intersection over Union (IoU).
The IoU value measures the overlap between the projected and predicted anomaly regions.
When calculating both the intersection and union area for IoU, the region that becomes null after projection is ignored in the calculations (usually because of occlusion or the region being too far in distance).
This IoU value indicates the level of consistency between the two masks, with higher values suggesting a greater degree of temporal consistency.

Formally, the metric is defined as:
\begin{align}
  f(\phi, \mathbf{X}, \mathbf{Y}) = \frac{1}{T - \Delta t} \sum_{t=1}^{T - \Delta t} \text{IoU}(\phi, \mathbf{x}_t, \mathbf{x}_{t+\Delta t}) \\
  \text{IoU}(\phi, \mathbf{x}_t, \mathbf{x}_{t + \Delta t}) = \frac{\# \{ P \wedge ( S_{t \rightarrow t + \Delta t} \wedge S_{t + \Delta t}) \} }{\# \{ P \wedge ( S_{t \rightarrow t + \Delta t} \vee S_{t + \Delta t}) \}}
\end{align}
where $T$ is the total number of frames in the video, and $\Delta t$ is the latency of the method.
$P$ is a binary mask denoting non-null regions after projection.
$\wedge$ and $\vee$ are pixel-wise logical intersection and union operations.
$\#\{ \cdot \}$ denotes the number of $1$ elements.
To ensure a fair comparison between methods with different latencies, we use a fixed value of $\Delta t = 1\text{s}$ for this metric.


\section{Benchmark Results}
\label{sec:Benchmark Results}

To validate the novelty of the proposed benchmark, we conduct a comprehensive evaluation of the state-of-the-art methods on the collected dataset.
The anomaly segmentation methods selected for evaluation can be splited into two categories: \textbf{(a)} methods that address the task without extra anomalous data or retraining of the network (e.g., SML\cite{jung2021standardized} and GMMSeg \cite{liang2022gmmseg}) \textbf{(b)} methods that leverage extra data with anomalous objects to retrain the segmentation networks (e.g., PEBAL\cite{tian2022pixel}, RPL\cite{liu2022residual} and DenseHybrid\cite{grcic2022densehybrid}).
During evaluation, we utilize an additional set of scenes without anomalous objects for the training of the semantic segmentation task, which is a preceding task of anomaly segmentation.
Of the 220 video sequences, 200 are used for training (if needed) and 20 are used for validation.


The results evaluated on the proposed metrics, i.e., the \textbf{latency-agnostic}, \textbf{latency-aware} metrics as well as the \textbf{temporal consistency}, are reported in Table.~\ref{table:main}.

From the results, we observe that the methods that leverage extra anomalous data to retrain the segmentation networks generally performs better than the methods that do not.
This observation is consistent with the intuition that the extra anomalous data can help the segmentation network to learn more discriminative features for anomaly segmentation.
However, we also discover that the temporal consistencies of the methods that leverage extra anomalous data and require retraining of the network are generally lower that the counter part.
This could be partly due to the network is fitted on the extra anomalous data and thus is less robust to the temporal perturbations.

Another observation is that the methods with higher performance on latency-agnostic metrics are more likely to be affected by the latency when evaluated on the latency-aware metrics.
Generally, performances with the latency-aware metrics are worse than those with the latency-agnostic metrics, as expected.
One exception is GMMSeg\cite{liang2022gmmseg} which shows similar performances under latency-agnostic and latency-aware metrics.
We attribute this phenomenon to the fact that, methods with higher latency-agnostic performances, are closer to the oracle performances demonstrated in Fig.~\ref{fig:oracle} and thus more sensitive to the latency.


\section{Summary}

In this work, we are motivated by a real-world scenario where anomaly segmentation methods are used to guide human intervention.
In such a scenario, both the accuracy and the latency of the anomaly segmentation methods are key factors for safe autonomous driving.

As no existing benchmark taking both factors into account during evaluation, we propose a novel benchmark for the task of road anomaly segmentation in a temporally informed setting.
The benchmark is composed of a large-scale video-based synthetic dataset, a publicly available toolkit to transfer the synthetic dataset to any given styles, and carefully designed latencty-aware and temporal consistency metrics preferring methods with both high accuracies and low latencites.
The overall design of the benchmark is to provide a well-established standard to measure the availability of the methods in the aforementioned scenario.

We hope that the proposed benchmark will encourage the development of new methods for road anomaly segmentation which would be applicable in the autonomous driving systems.

\newpage

{
    \small
    \bibliographystyle{ieeenat_fullname}
    \bibliography{main}
}

\end{document}